\documentclass[10pt]{article}

\usepackage[utf8]{inputenc}
\usepackage[margin=0.8in]{geometry}
\usepackage{graphicx}
\usepackage{amsmath}
\usepackage{amssymb}
\usepackage[superscript,biblabel]{cite}
\usepackage{algorithm}
\usepackage{algpseudocode}
\usepackage{tikz} 
\usetikzlibrary{shapes.geometric, arrows}
\usepackage{caption}
\usepackage{booktabs}
\usepackage[affil-it]{authblk}

\renewcommand{\thefootnote}{\roman{footnote}}

\newcommand\blfootnote[1]{%
	\begingroup
	\renewcommand\thefootnote{}\footnote{#1}%
	\addtocounter{footnote}{-1}%
	\endgroup
}

\usepackage{setspace}
\tikzstyle{startstop} = [rectangle, rounded corners, minimum width=3cm, minimum height=1cm,text centered, draw=black, fill=red!30]
\tikzstyle{io} = [trapezium, trapezium left angle=80, trapezium right angle=100, minimum width=3cm, minimum height=1cm, text centered, text width=3cm, draw=black, fill=purple!30]
\tikzstyle{Rprocess} = [rectangle, minimum width=3cm, minimum height=1cm, text centered, text width=3cm, draw=black, fill=blue!30]
\tikzstyle{MATprocess} = [rectangle, minimum width=3cm, minimum height=1cm, text centered, text width=3cm, draw=black, fill=orange!30]
\tikzstyle{LASprocess} = [rectangle, minimum width=3cm, minimum height=1cm, text centered, text width=3cm, draw=black, fill=green!30]
\tikzstyle{decision} = [diamond, minimum width=2cm, minimum height=1cm, text centered, text width=2cm, draw=black, fill=green!30]
\tikzstyle{arrow} = [thick,->,>=stealth]

\begin{document}
\title{Three-dimensional Segmentation of Trees Through a Flexible Multi-Class Graph Cut Algorithm (MCGC) }

\author[a,b,c,*]{Jonathan Williams}
\author[b]{Carola-Bibiane Sch\"onlieb}
\author[a,c,d]{Tom Swinfield}
\author[b]{Juheon Lee}
\author[b]{Xiaohao Cai}
\author[e]{Lan Qie}
\author[a,c]{David A. Coomes}

\affil[a]{Forest Ecology and Conservation Group, Department of Plant Science, University of Cambridge, CB2 3EA, UK}
\affil[b]{Image Analysis Group, Department of Applied Mathematics and Theoretical Physics (DAMTP), University of Cambridge, CB3 0WA, UK}
\affil[c]{University of Cambridge Conservation Research Institute (UCCRI), David Attenborough Building, Cambridge, CB2 3QY, UK}
\affil[d]{Centre for Conservation Science, Royal Society for Protection of Birds, David Attenborough Building, Cambridge, CB2 3QY, UK}
\affil[e]{School of Life Sciences, University of Lincoln, Brayford Pool, Lincoln, LN6 7TS, UK}
\affil[*]{Corresponding author: jonvw28@gmail.com}

\maketitle

\begin{abstract}

\blfootnote{Preprint submitted to IEEE. \copyright 2019 IEEE.  Personal use of this material is permitted.  Permission from IEEE must be obtained for all other uses, in any current or future media, including reprinting/republishing this material for advertising or promotional purposes, creating new collective works, for resale or redistribution to servers or lists, or reuse of any copyrighted component of this work in other works.}

Developing a robust algorithm for automatic individual tree crown (ITC) detection from laser scanning datasets is important for tracking the responses of trees to anthropogenic change. Such approaches allow the size, growth and mortality of individual trees to be measured, enabling forest carbon stocks and dynamics to be tracked and understood. Many algorithms exist for structurally simple forests including coniferous forests and plantations. Finding a robust solution for structurally complex, species-rich tropical forests remains a challenge; existing segmentation algorithms often perform less well than simple area-based approaches when estimating plot-level biomass. Here we describe a Multi-Class Graph Cut (MCGC) approach to tree crown delineation. This uses local three-dimensional geometry and density information, alongside knowledge of crown allometries, to segment individual tree crowns from LiDAR point clouds. Our approach robustly identifies trees in the top and intermediate layers of the canopy, but cannot recognise small trees. From these three-dimensional crowns, we are able to measure individual tree biomass. Comparing these estimates to those from permanent inventory plots, our algorithm is able to produce robust estimates of hectare-scale carbon density, demonstrating the power of ITC approaches in monitoring forests. The flexibility of our method to add additional dimensions of information, such as spectral reflectance, make this approach an obvious avenue for future development and extension to other sources of three-dimensional data, such as structure from motion datasets.
\end{abstract}

\section{Introduction}
\label{sec:intro}

Automatically identifying and mapping trees is a long-standing goal within the field of forest remote sensing, and there is currently particular interest in finding robust solutions for segmenting multi-species stands with complex structures \cite{larsen2011,Kaartinen2012,Dalponte2014,Eysn2015,Zhen2016,Lindberg2017}. With the widespread adoption of Light Detection and Ranging (LiDAR) it is possible to collect data on the three-dimensional (3D) structure of forest stands over hundreds of square kilometers in a matter of hours, commonly in the form of 3D point clouds \cite{Xie2008,Petrou2015, toth2016}, from which individual trees can be segmented. Individual tree approaches have the potential to track individual-level changes in response to events such as disease, biological invasion, logging and extreme climatic events (e.g. droughts) thereby gaining a clearer understanding of the processes that generate structural change\cite{Asner2008,Andersen2014,vanEwijk2014,Barnes2017}.

Until recently, most approaches to delineating individual tree crowns (ITCs) relied on converting the 3D structural information contained in LiDAR datasets into rasterised 2D surface models\cite{larsen2011,Kaartinen2012,Zhen2016}. Typically, these approaches start by finding local maxima which are treated as tree tops, then searching around those peaks to find tree crowns by methods such as the watershed algorithm\cite{Chen2006,Koch2006,Zhao2014,Liu2015}, region growing\cite{Solberg2006,Zhen2015,Dalponte2016}, valley-following\cite{Gougeon1995,Leckie2005,Katoh2012} or variable-window filtering\cite{Wulder2000,Hyyppa2001}. Whilst these methods are often very successful at finding the largest trees, particularly in coniferous forests where trees have clearly defined tops, conversion to 2D rasters results in the loss of almost all the information from the understory, including many smaller trees, which are represented in the full 3D point cloud \cite{Zhen2016,Dalponte2016,Lindberg2017,Hamraz2017}. For applications such as forest type and cover mapping and above-ground biomass (AGB) estimation these methods can deliver robust results, but for estimating tree size distributions or tree-centric changes this loss of data is problematic \cite{Coomes2017}.

Segmentation approaches that make better use of LiDAR point clouds are becoming more prevalent\cite{Zhen2016,Lindberg2017}. The earliest methods worked with voxels: data summarised within 3D grids rather than 2D rasters,\cite{Popescu2008,Wang2008,Wu2013} but were highly affected by variation in point density\cite{Zhen2016}. More recently, methods have emerged which produce a full 3D version of the tree by taking points directly from the point cloud using minimal or no summarisation\cite{Lindberg2017}. Other early approaches made use of common clustering algorithms, such as k-means, to find groups of points thought to be trees\cite{Morsdorf2003,Gupta2010,Lindberg2014}. These produce reasonable results for larger trees, especially in simpler conifer forests where crowns are typically distinct\cite{Kaartinen2012}. Such methods are less reliable at finding smaller trees or delineating structurally complex stands\cite{Lindberg2017}, such as tropical rain forest and broadleaf temperate forest. These are challenging forests to segment because trees often have overlapping crowns\cite{Vauhkonen2012} and stands contain multiple layers of trees\cite{Eysn2015}. To address the multiple layered complexity, methods have been developed which apply an algorithm more than once to point-cloud data, locating and segmenting tall trees, which are then removed from the point cloud before reiterating the algorithm \cite{Vega2011,Duncanson2014}. The latest methods combine a clustering approach with constraints based on the allometry of trees: knowledge of how the size and shape of crowns changes with tree height within a particular forest type \cite{Jing2012,Hu2014}. For example, \textit{AMS3D} combines a locally derived allometric relationship between height and crown diameter with an adaptive mean shift algorithm\cite{Ferraz2012,Ferraz2016a,Ferraz2016b} and \textit{Ptrees} combines a multi-scale nearest-neighbour algorithm with analysis of crown geometry \cite{Vega2014}. Finally, normalised cut methods, from graph theory, have been applied by using recursive binary cuts (i.e. repeated division of segments into two components), until some stopping point is reached\cite{Reitberger2009,Lee2017}. Because this approach works with a matrix of distances between points in the cloud, it is computationally intensive, and is only applied as a refinement step \cite{Reitberger2009,Yao2013,Yu2015,Yang2016,Zhong2017} or applied to voxelised datasets \cite{Reitberger2009,Yao2012,Yao2013}. Only recently has the use of a Multi-Class Graph Cut approach been applied to the problem of ITC detection to data from European coniferous forests, focusing on Terrestrial LiDAR data and working with voxelised data \cite{Heinzel2018}.

This paper aims to overcome a number of limitations to current segmentation algorithms working with LiDAR point clouds. The first issue is that algorithms are commonly trained with allometric data from specific forest types and adjusting them to other regions can be difficult\cite{Lindberg2014}. Secondly, they are developed to work on topographically corrected data, where the ground height is subtracted from the heights of non-ground returns; this process warps tree crown geometries when the ground is not level (but see [\citen{Vega2014}]). Finally, most methods work with spatial information alone, and are not designed to include spectral information; hyperspectral or multi-spectral imagery collected alongside the LIDAR datasets \cite{Petrou2015, Zhen2016}. Here we develop a novel and flexible approach based upon a Multi-Class Normalised Graph Cut which works directly on the raw point cloud\cite{shi2000} and addresses all of these issues. Our method converts the point cloud into a graph representation, with a vertex for each pixel. Proximity of vertices in the graph represents closely related pixels. This conversion is based on 3D proximity as well as assessment of the local density of the point cloud. Four parameters allow tuning of the importance of vertical and planimetric distances and vertical and horizontal variations in density to different forest types. We describe how this approach can be extended to include spectral differences in the linkage calculation, though this is not explored in this study. The algorithm then simultaneously splits the graph into clusters of points which are well-connected. The number of clusters is automatically determined based on the structure of the graph produced from the data, and these result in candidate tree crowns. Through a filtering step, using an allometric crown-geometry relationship derived from a regional dataset, the clusters are checked for feasibility and rejected if they don't meet a set of size criteria or contain too few points. From this a set of tree crowns is produced, along with rejected points not yet assigned to a crown. Through a second application of the algorithm it is then possible to address some of the issue of multi-layer canopies. We find that the difficulties of finding obscured lower canopy trees from ALS data remains with our method, but that a double-layer MCGC approach produces a good proxy for field inventory of stems in all but the lowest canopy layers.

We also look at addressing the computation of biomass at a plot level. We use MCGC combined with a relationship for tree biomass from crown measurements to predict total plot biomass as a sum of that contributed by each tree. Many analyses currently focus on summarised data at the plot level, computing metrics such as the distribution of top of canopy height and variation in canopy structure and using these to predict useful forestry attributes such as biomass or canopy fuel mass\cite{Drake2003,Andersen2005,Asner2010,Asner2012}. These give good estimates of biomass and carbon density, but lose information on the individual trees found in each plot\cite{Coomes2017,Nunes2017}. Previous work suggests that these methods are preferable to ITC approaches\cite{Coomes2017} but we show that the MCGC approach to ITC is able to produce improvements in estimations of hectare scale biomass. This advancement shows the potential of ITC approaches in forest management.

\section{Tree Crown Segmentation with Multi-Class Graph Cut}
\label{sec:frame}

This section outlines the mathematics underpinning a graph cut, summarising the formal work already completed on this \cite{shi2000,vonLuxburg2007,Hastie2009}. A \textit{graph}, $G = G(V,E)$, is a coupled pair of a set of vertices, $V$, and edges, $E \subset V \times V$, which connect these vertices. Each edge has an associated non-negative weight, $w_{ij} \geq 0$, which represents how closely related the vertices $\{v_{i},v_{j}\}$ are, with larger values representing stronger associations. In this work we use undirected graphs, so $w_{ij} = w_{ji}$. The choice of the weights is application-dependent and indeed crucial for the performance of the graph cut approach. For the tree crown segmentation algorithm proposed in this paper the choice of the weights will be discussed in Section \ref{ssubsec:meth_tcs_gcalg}. 

A \textit{Binary Cut} of a graph into two disconnected subsets ($A$ and $B$, forming a partition of $G$) has an associated cost defined as $\mathrm{cut}(A,B) = \sum_{i, j: v_{i} \in A,\hspace{0.2em} v_{j} \in B} w_{ij}$, which is the sum of weights for all links that bridge the cut. Minimising this cut over possible partitions A and B means that we are looking for a partition which maximises the dissimiliariy between the two sets. Similarly a \textit{Multi-Class Graph Cut} of $G$ into $k$ subsets $\{A_{i}\}_{i \in 1,...,k}$ forming a partition of $G$ has an associated cost defined as:

\begin{equation}
\mathrm{cut}(A_{1},...,A_{k}) = \frac{1}{2} \sum_{i=1}^{k} \mathrm{cut}(A_{i},\overline{A}_{i}),
\label{eq:multicut}
\end{equation}

\noindent
where $\overline{A}_{i}$ is the complement of $A_{i}$, so that $\overline{A}_{i} = \{v_{j} | v_{j} \notin A_{i}\}$. Finding a partition $\{A_{i}\}_{i \in 1,...,k}$ that minimises \eqref{eq:multicut} can lead to qualitatively very good segmentations, but also tends to favour cutting small sets of isolated vertices \cite{wu1993optimal}. Indeed, without additional constraints, a partition where the whole of the graph is contained in one subset and all other subsets are empty is optimal for \eqref{eq:multicut}. To overcome this a balanced graph cut is used. This splits the graph into similarly sized subsets (defined by their volume) whilst also trying to minimise linkages between these sets. Here the \textit{Normalised Multi-Class Graph Cut} \cite{shi2000} is used with an associated cost:

\begin{equation}
\mathrm{Ncut}(A_{1},...,A_{k}) = \frac{1}{2} \sum_{i=1}^{k} \frac{\mathrm{cut}(A_{i},\overline{A}_{i})}{\mathrm{vol}(A_{i})},
\label{eq:NMulticlass}
\end{equation}
where $\mathrm{vol}(A)$  is the \textit{volume} of a set of vertices $A\subset V$ and defined as $\mathrm{vol}(A) = \sum_{i,j: v_{i} \in A, v_{j} \in V} w_{ij}$. The volume quantifies sets with highly globally connected vertices as `larger' than poorly connected sets.
This formulation of the problem enforces segmentation into subsets of more balanced size and reduces the chances of a single dominant cluster, in line with the problem of tree crown segmentation. The trade-off is that it becomes a discrete optimization problem which is NP-complete and difficult to solve directly\cite{shi2000,vonLuxburg2007}. To overcome this, an approximation based on trace minimisation of a matrix is used. This is a well-studied problem, and can be solved by spectral clustering methods\cite{vonLuxburg2007}. The trace minimisation problem that approximates the minimisation problem for Normalised Multi-Class Graph Cut \eqref{eq:NMulticlass} is:

\begin{equation}
\mathrm{Relaxed \hspace{0.2em} Normalised \hspace{0.2em} Cut} : \underset{T \in \mathbb{R}^{n \times k}}{\mathrm{min}} \mathrm{Tr}(T^{\top}D^{-\frac{1}{2}}(D - W)D^{\frac{1}{2}}T) \hspace{0.5em} \mathrm{subject \hspace{0.2em} to} \hspace{0.5em} T^{\top}T = I
\label{eq:RelaxNMulticlass}
\end{equation}.

\noindent
where $W = (w_{ij})_{i,j}$ is the matrix of pairwise weights between vertices $v_i$ and $v_j$ and $D=(d_{ij})_{i,j}$ is the diagonal matrix of degrees for each vertex, that is $d_{ii} = \sum_{j:v_{j} \in V} w_{ij}$ and $d_{ij}=0$ for $i\neq j$. Moreover, $I$ is the $k \times k$ identity matrix and $T$ is the $n \times k$ matrix used as the basis for spectral clustering, where each row $i$ represents one of the $n$ vertices $v_i$. Problem \eqref{eq:RelaxNMulticlass} is then solved by Normalised Spectral Clustering as per Algorithm \ref{alg:normspec}:

\begin{algorithm}
	\caption{Normalised Spectral Clustering [\citen{shi2000}]}
	\label{alg:normspec}
	\begin{algorithmic}	[1]
		\State{Compute the normalised graph Laplacian $L_{\mathrm{sym}} = D^{-\frac{1}{2}}(D-W)D^{\frac{1}{2}}$}
		\State{Compute the first $k$ eigenvectors $u_{1},\cdots,u_{k}$ of $L_{\mathrm{sym}}$}
        \State{Form the matrix $T$ $\in \mathbb{R}^{n \times k}$ from $u_1,\cdots,u_k$ by setting $t_{ij} = u_{ij}/(\sum_{k}u_{ik}^{2})^{1/2}$}
		\State{Cluster the rows of $T$ with $k$-means clustering\cite{Hastie2009} into clusters $C_{1},...,C_{k}$}
		\State{Return clusters $A_{1},...,A_{k}$ with $A_{i} = \{j\in V~|\,\mathrm{row}\, j \mathrm{\,of\, T} \in C_{i}\}$}
	\end{algorithmic}
\end{algorithm}

In previous studies, graph-cut based segmentation of tree crowns has been used with recursive binary cuts, or with a Multi-Class Cut with a predetermined number of clusters $k$ \cite{Reitberger2009,Lee2015b,Lee2016,Yang2016,Lee2017}. In this paper, we instead adapt the approach in \cite{Little2015} and allow $k$ to vary. We propose the \textbf{Multi-Class Graph Cut (MCGC)} approach where a minimum and maximum number of clusters are set and the eigenvectors up to the maximum number of clusters are computed. For associated eigenvalues $\lambda_{i}$ (where $\lambda_{1}$ is the smallest eigenvalue) the \textit{eigengaps} $\lambda_{i+1} - \lambda_{i}$ are computed. These represent the relative stability of computed segmentations, with bigger gaps for greater variation when extending segmentation to include additional groups. Hence choosing a segmentation which avoids a large change in variation produces a more stable segmentation. \cite{Azran2006,vonLuxburg2007,Kong2010,Little2015}. The number of clusters is then set by the eigenvalue $\lambda_{i}$ which maximises the eigengap $\lambda_{i+1} - \lambda_{i}$ within the pre-defined minimum and maximum range, allowing flexibility based upon the data.

\section{Methods}
\label{sec:meth}

\subsection{Field datasets}
\label{subsec:meth_data}

Data for this paper were taken from a lowland tropical rainforest inventory reserve in Sabah, Malaysia. Sepilok reserve is a 4294 ha protected area within which three distinctive forest types are found: open 80 m tall alluvial forest, dense 60 m tall sandstone hill forest and 30 m tall Kerangas forests on the shallow soils of hill tops. The elevation in this lowland forest ranges from 0 to 250 m a.s.l.\cite{Coomes2017}.

\textbf{LiDAR:} The reserve was surveyed with ALS data on 5th November 2014, using a Leica ALS50-II ALS flown at 1850 m altitude on-board a Dornier 228-201 travelling at 135 knots. The sensor emitted pulses as 83.1 Hz with a field of view of $12.0^{\circ}$ and footprint of approximately 40 cm diameter with an average pulse density of 7.3 points per $\mathrm{m}^{2}$. A total area of 26 $\mathrm{km}^2$ was covered, including all 9 permanent plots below. The sensor records full waveform ALS, but for this study the data were discretised by the system, with up to four returns per pulse. A nearby Leica base station was used to ensure accurate georeferencing of the data. The data were pre-processed by NERC's Data Analysis Node and delivered in standard LAS format.\cite{Coomes2017} 

\textbf{Permanent plots:} In order to validate the segmentation approach, we compared the outputs with field-measurements. Three permanent inventory plots of $200 \times 200$ m are established in each of the three soil types which have been regularly surveyed through out their history; the most recent survey, used for this analysis, was completed in 2013--15. Field inventory data included the diameters of all stems $\geq$ 5 cm in diameter (measured at a height of 1.3 m) and their species identity, mapped to the nearest 10 $\times$ 10 m subplot. For some stems, a field-measured height was also recorded. Additionally, in [\citen{Coomes2017}], a set of 91 crowns were identified in the LiDAR dataset and subsequently mapped in the field. These include field and LiDAR based measures of height, stem diameter and crown area.

\textbf{Allometry:} Having allometric relationships that estimate crown size for trees of a given size is important for accurate segmentation, for reasons explained below. Allometric relationships were obtained by subsetting the Indo-Malaya region from a global allometry dataset \cite{Jucker2017}. In total 7,943 trees were included, ranging in height from 1.4 m to 70.7 m. Quantile log-log regression (from the \texttt{quantreg} R package)\cite{R2016} was used to fit `median' and `upper boundary' relationships through the data, with the following form: $CD = \alpha H^{\beta}$ where $\mathrm{CD}$ is crown diameter, $\mathrm{H}$ is tree height with both measured in metres and $\alpha$ and $\beta$ are fitted. The 50th and 95th percentiles of this relationship were computed as follows:

\begin{equation}
\begin{aligned}
\mathrm{CD}_{\mathrm{IM_{50}}} & = 0.251 \times \mathrm{H}^{0.830}, \\
\mathrm{CD}_{\mathrm{IM_{95}}} & = 0.446 \times \mathrm{H}^{0.854}.
\label{eq:allom_H_CD_TF_me}
\end{aligned}
\end{equation}These relationships were then used to predict crown diameter and radius for trees of each height when applying allometry, as explained in the next section.

The field inventory at Sepilok includes stem diameter but rarely records tree height as needed to estimate tree biomass as explained in Section \ref{ssec:meth_biomass}. To compute the height of trees in the field inventory at Sepilok we used the relationship described by Coomes et al. in [\citen{Coomes2017}]. Here height, $\mathrm{H}$ (m), is predicted as a function of stem diameter, $\mathrm{D}$ (cm), as a power law, with a different relationship applied to each soil type as follows:

\begin{equation}
\begin{aligned}
\mathrm{H}_{\mathrm{Alluvial}} & = 2.105 \times \mathrm{D}^{0.679}, \\
\mathrm{H}_{\mathrm{Kerangas}} & = 4.57 \times \mathrm{D}^{0.461}, \\
\mathrm{H}_{\mathrm{Sandstone}} & = 4.001 \times \mathrm{D}^{0.527}.
\label{eq:allom_H_D}
\end{aligned}
\end{equation}
\subsection{The Multi-Class Graph Cut algorithm}
\label{subsec:meth_tcs}

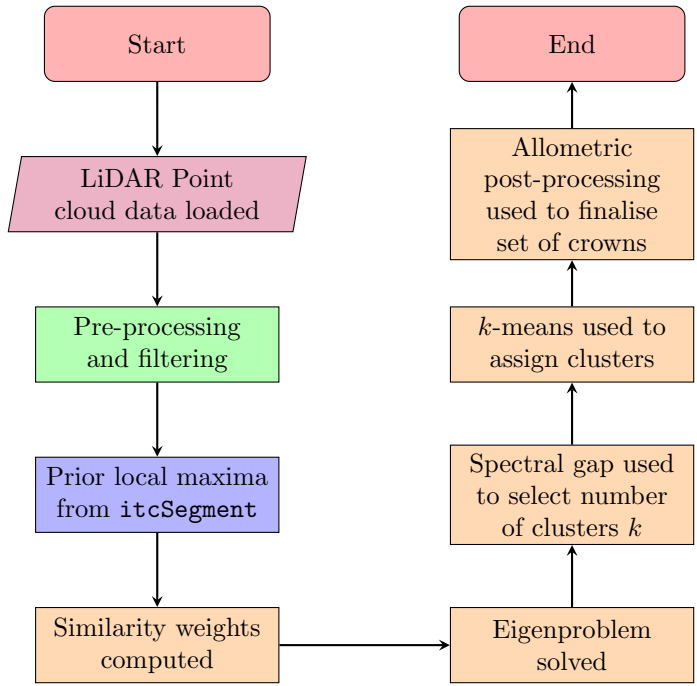
\begin{figure}[!t]
	\centering
	\begin{tikzpicture}[node distance=2cm]
	\node (start) [startstop] {Start};
	\node (in1) [io, below of=start] {LiDAR Point cloud data loaded};
	\node (pro1) [LASprocess, below of=in1] {Pre-processing and filtering};
	\node (pro2) [Rprocess, below of=pro1] {Prior local maxima from \texttt{itcSegment}};
	\node (pro3) [MATprocess, below of=pro2] {Similarity weights computed};
	\node (pro4) [MATprocess, right of=pro3, xshift = 10em] {Eigenproblem solved};
	\node (pro5) [MATprocess, above of=pro4] {Spectral gap used to select number of clusters $k$};
	\node (pro6) [MATprocess, above of=pro5] {$k$-means used to assign clusters};
	\node (pro7) [MATprocess, above of=pro6] {Allometric post-processing used to finalise set of crowns};
	\node (end) [startstop, above of=pro7] {End};
	
	\draw [arrow] (start) -- (in1);
	\draw [arrow] (in1) -- (pro1);
	\draw [arrow] (pro1) -- (pro2);
	\draw [arrow] (pro2) -- (pro3);
	\draw [arrow] (pro3) -- (pro4);
	\draw [arrow] (pro4) -- (pro5);
	\draw [arrow] (pro5) -- (pro6);
    \draw [arrow] (pro6) -- (pro7);
	\draw [arrow] (pro7) -- (end);
	\end{tikzpicture}
	\caption{MCGC tree crown segmentation algorithm. Here the colour of processes reflects the environment they have been coded in: LASTools are marked in green, R code is marked in blue and MATLAB code is coloured in orange.}
	\label{fig:gc_fullalg}
\end{figure}

 The MCGC approach, summarised in Figure \ref{fig:gc_fullalg}, is explained in detail in this section. LAS data is first pre-processed and then the MCGC algorithm is applied to delineate tree crowns. Further post-processing based on knowledge of tree architectural geometry (i.e. allometric relationships) is then applied to ensure only sensibly shaped crowns are retained. A final double-layer extension is then explained.

\subsubsection{Pre-processing and prior generation}
\label{ssubsec:meth_tcs_preproc}

The point cloud data are first cleaned using the LASTools package\footnote{https://rapidlasso.com/lastools/}. Points marked as noise are removed and the \texttt{lasheight} method is used to generate a second point cloud with a model of the ground subtracted from the data points (based on points labeled as ground) to enable computation of above-ground height of points for prior generation.

The accuracy of the graph cut is improved by generating a lower bound on the number of expected tree crowns. This is estimated using the topography-corrected height point cloud and tree allometry information. Here we produced a list of expected tree-top positions computed by itcSegment --- a local maximum finder that scales its search window size in relation to tree height used in the R package \texttt{itcSegment}\cite{R2016,Coomes2017}. Work presented here is completed using a rasterised canopy height model (CHM) gridded to 0.5 $\times$ 0.5 m pixels. A 50th percentile relationship between height and crown diameter was used to set the size of the local search window based on the height of the CHM at that location. In this study the relevant relationship in equation \eqref{eq:allom_H_CD_TF_me} was always used in computing the prior. The number of local maxima found then sets a lower bound for the total number of trees to be found by MCGC. As itcSegment only works on the top of canopy, it is expected to miss many lower canopy trees and so this number should be an underestimate. The MCGC approach, as outlined in Section \ref{sec:frame}, requires a minimum and maximum number of tree (clusters) per plot to work with. The minimum matching the number of prior `trees' and a maximum of twice this value was set. This choice was made to avoid forcing over splitting of crowns in the top, dominant layer of the canopy. We justify the suitability of this choice in the discussion in Section \ref{ssubsec:dis_MCGC_FI}.

\subsubsection{Graph cut}
\label{ssubsec:meth_tcs_gcalg}

\begin{figure}[t!]
\begin{center}
\includegraphics[width=1\textwidth]{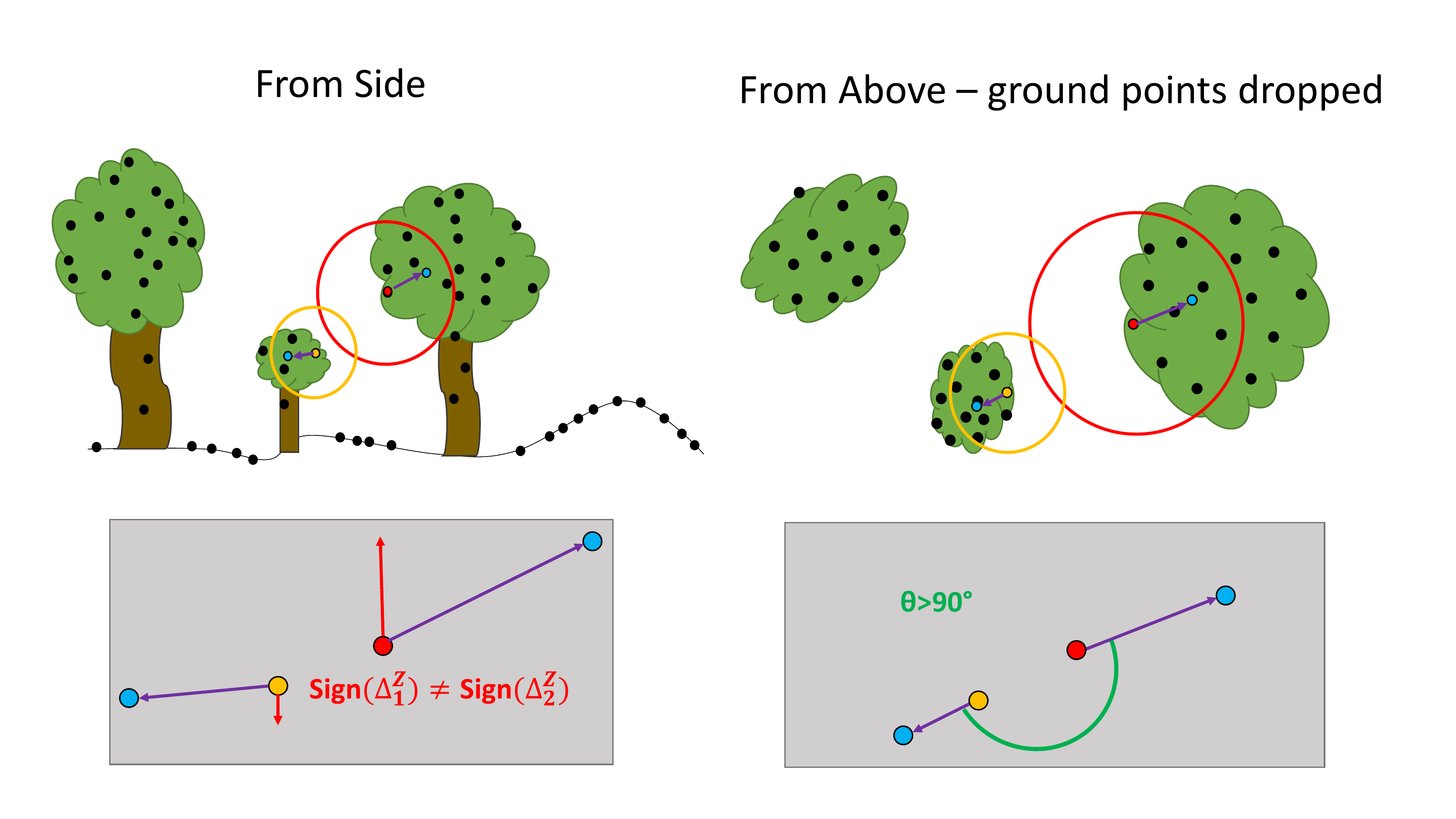}
\caption{Simplified illustration of the principle of the local density centroid calculations. Points being compared are red and orange, with associated neighbourhoods shown. Blue points are centroids for each neighbourhood, with centroid vectors in purple. Comparisons of these vectors are then shown in the grey boxes. In this example both comparisons would result in a reduction in the similarity weight between the highlighted points. Here the illustration shows how the horizontal and vertical directional differences in the centroid vectors distinguish the points highlighted as belonging to distinct crowns.}
\label{fig:geom_diagram}
\end{center}
\end{figure}

\begin{figure}[t!]
\begin{center}
\includegraphics[width=1\textwidth]{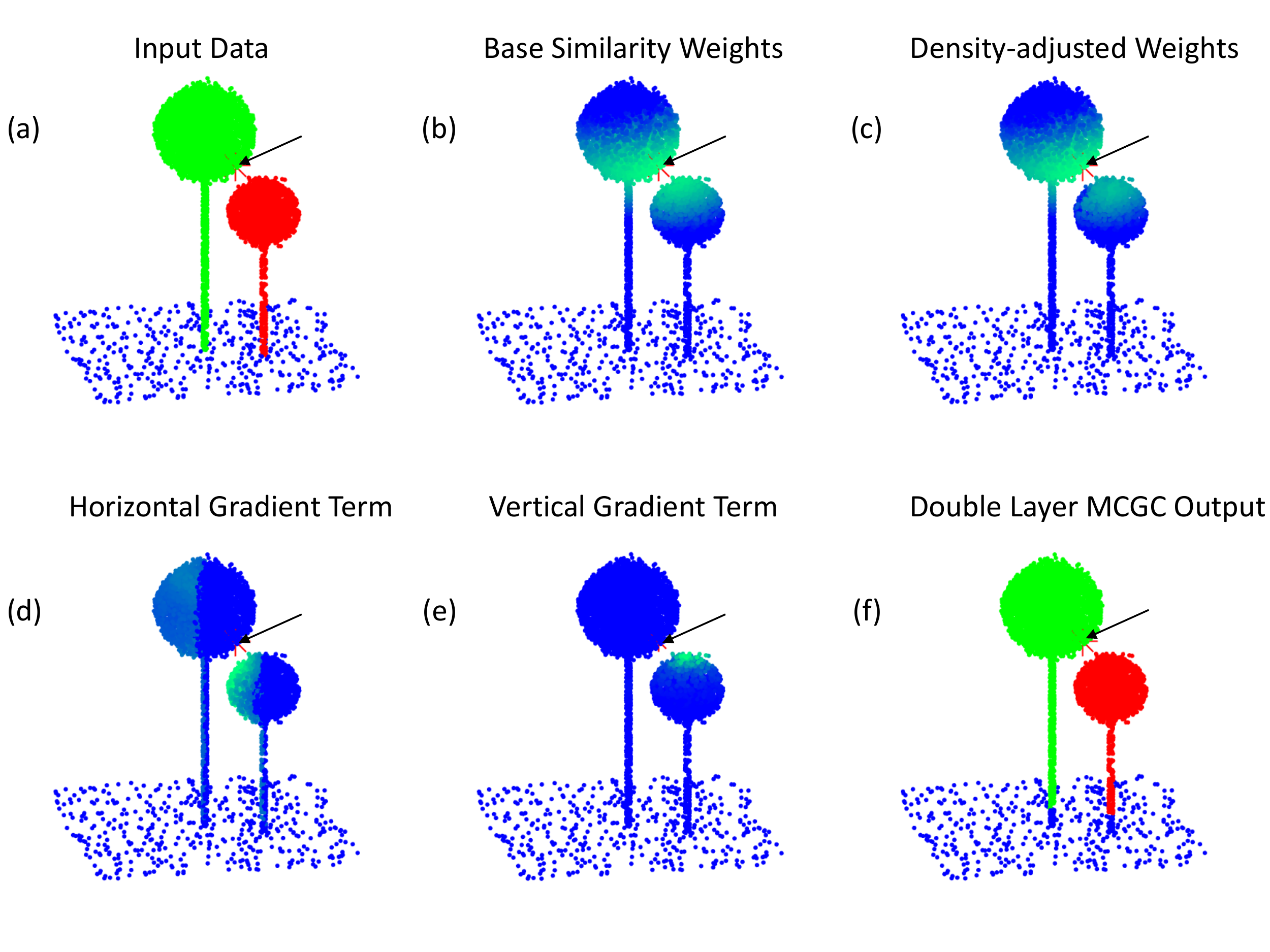}
\caption{MCGC applied to artificial trees to explain the role of centroid vector computations in segmentation. The relative effect of applying the centroid adjustments when comparing to the marked point are shown in panels (d) and (e), and this is reflected in the weights as shown in panels (b) and (c), where the closest points on the smaller crown have weaker links than the base weights, with minimal difference made to points in the taller crown. Panel (a) shows two simulated trees with allometry matching that of the 50th percentile for crowns in Indo-malaya where the green tree is 30 m tall and red tree is 20 m tall with blue points being the ground. Panel (b) shows the qualitative distribution of $w_{ij}^{\mathrm{base}}$ for each point computed with reference to the crown boundary point highlighted by an arrow in all panels, centred at the red star; higher values are represented by green in the blue-green colour ramp. Panel (c) shows the distribution of $w_{ij}$ once the centroid adjustments have been made. The colour ramp is the same and shows the reduction in similarity to the neighbouring crown whilst preserving linkages to the target crown. (d) shows the effect of the modification based on horizontal components of centroid vectors in equation \eqref{eq:w_H}. Dark blue represents minimal adjustment to the weight and lighter colours represent a greater reduction in weight. (e) shows the same effect as in (d) when looking at the adjustment based on the vertical component of the centroid vectors in equation \eqref{eq:w_Z}. (f) shows the output of double-layer MCGC, with each crown highlighted in a different colour, with unclassified points in blue, recovering the true structure in (a) with the exception of the bottom 2 m of the stem, as this was the height threshold set.}
\label{fig:geom_ex}
\end{center}
\end{figure}

In what follows, we explain in detail the steps involved in our application of the MCGC graph cut algorithm introduced in Section \ref{sec:frame}. The first step of the graph cut algorithm (implemented in MATLAB) converts 3D co-ordinates into a graph representation. Here the raw height above sea level values are used to preserve the true structure of the vegetation. Working with topographically corrected data causes crown structures to be warped when the terrain is not flat as points are measured from the ground directly below them and not from the base of their respective tree stem\cite{Vega2014}. In this work the 3D information in the LiDAR point cloud was used to construct weights to represent the similarity between points. The performance of graph cut approaches is contingent on the choice of weights, $w_{ij}$, and these must reflect the structure of segmentation targets to produce good results. Here we introduce a novel approach to assessing the similarity of points in a 3D point cloud for tree detection which avoids the need for prior determination of tree tops or stem locations and only uses the information within the structure of the point cloud. Our process captures knowledge of the basic geometry of the data and of the local density of points and how this relates to tree crowns. The key idea in using the local density of points is that for points near the external boundary of a crown, the local density of points constituting the crown should be higher than that of points outside the crown. Computing a centre of mass (centroid) for the points in a neighbourhood of a boundary point should result in a point which lies roughly between the centre of the crown and the boundary point on which the neighbourhood is centred. Two nearby points on the boundary of the same crown should have their local centroids in a similar direction relative to the respective boundary points. In contrast, two boundary points on different crowns will have their local centroids located towards the middle of their respective crowns and so relative to the two boundary points, these will be orientated in different directions. Comparing the relative orientation of local centroids with respect to points being compared should help distinguish crown boundaries. A simplified example of this principle is illustrated in Figure \ref{fig:geom_diagram} and a practical implementation on artificial data is shown in Figure \ref{fig:geom_ex}.

To encode the information from the local density of points into the similarity weights, the local centroid for each point was computed. The neighbourhood used was a sphere centred on the point in question with a radius based upon the above-ground height of the point in question. The radius of the sphere was based on the 95th percentile relationship in equation \eqref{eq:allom_H_CD_TF_me}. To avoid the sphere extending beyond the stem location for boundary points in most trees the radius of the sphere was set to be half of the reference radius computed from the allometric relationship. This would capture most of the points in the relevant half of crown without extending too far and being skewed by other trees locally. The centroid of all points in this neighbourhood and the relative vector from the original point to its centroid, denoted by $\Delta_i$ for point $i$, was then computed. In comparing points, the orientations of these relative vectors were compared. Points which belong to boundaries of different crowns would be expected to have vectors with very different orientations. Each of these vectors was broken down into horizontal and vertical components, denoted $\Delta_{i}^H$ and $\Delta_{i}^Z$ respectively. Where the orientations suggested membership of different crowns, the similarity value ($w_{ij}$) was reduced. The process of computation of $w_{ij}$ is outlined as follows (with $w_{ii}=0, \hspace{0.5em} \forall i$). Note that it would be straightforward to introduce further similarity comparisons when constructing weights, such as comparing intensity values in colour channels from imagery, but we have not done so here.

First, a basic similarity based on the distance between each pair of points is computed as
\begin{equation}
w_{ij}^{\mathrm{base}} = \mathrm{exp}\bigg(- \frac{||(x,y)_{i}-(x,y)_{j}||^{2}_{2}}{\sigma_{\mathrm{XY}} ^{2}}\bigg) \times \mathrm{exp}\bigg(- \frac{(z_{i}-z_{j})^{2}}{\sigma_{\mathrm{Z}} ^{2}}\bigg).
\label{eq:base_w}
\end{equation}
Here, $w_{ij}^{\mathrm{base}}$ in equation is separated into a horizontal (first exponential term in \eqref{eq:base_w}) and a vertical component (second exponential in \eqref{eq:base_w}). Each component can have its importance controlled by separate parameters as the vertical and horizontal structure and extent of crowns can vary for different forest types. The horizontal and vertical parameters are set before applying MCGC and are the same for all points ($\sigma_{xy}$ and $\sigma_{Z}$ respectively). Here $(x,y,z)_{i}$ are the raw coordinates of point $i$. An example of this is illustrated in Figure \ref{fig:geom_ex}b.

Next, the horizontal angle between the two centroid vectors $\Delta_{i}^\mathrm{H}$ and $\Delta_{j}^\mathrm{H}$ is computed as $\theta_{\mathrm{H}}(i,j)$. When this is more than a right-angle, the basic similarity $w_{ij}^{\mathrm{base}}$ is reduced. This reduction is based on the horizontal distance between the points, $\mathrm{d}_{\mathrm{H}}$, so that closer points get a larger reduction in similarity. This effect is normalised by a scale parameter, $K_{\mathrm{H}}$, which is automatically set to the allometric radius for the tallest point in the data, using the 95th percentile allometry relationship in equation \eqref{eq:allom_H_CD_TF_me}. Similarly, the larger the relative difference in the horizontal components of the centroid vectors, the larger the reduction in the similarity. The parameter $W_{\mathrm{H}}$ controls the overall importance of this modification relative to all other steps in the computation. An example of this reduction is shown in Figure \ref{fig:geom_ex}e. With this, the basic $w_{ij}^{\mathrm{base}}$ is updated to $w_{ij}^{\mathrm{postH}}$ as

\begin{equation}
w_{ij}^{\mathrm{postH}} = 
\begin{cases}
w_{ij}^{\mathrm{base}} \times \mathrm{exp}\bigg(-W_{\mathrm{H}} \times \frac{K_{\mathrm{H}}}{\mathrm{d}_{\mathrm{H}}}\times||\Delta_{i}^\mathrm{H}-\Delta_{j}^\mathrm{H}||_{_{2}}\bigg), & \text{if } \theta_{\mathrm{H}}(i,j) > \frac{\pi}{2}, \\
w_{ij}^{\mathrm{base}},&\mathrm{otherwise}.
\end{cases}
\label{eq:w_H}
\end{equation}

Finally, the vertical components of the centroid vectors are compared ($\Delta_{i}^\mathrm{Z}$ and $\Delta_{j}^\mathrm{Z}$). When these point in the same direction, no adjustment is made to the score. Where the directions differ, only pairs where these diverge have their weight reduced. Divergence occurs when the taller of the points has a positive $\Delta Z$ and the lower point a negative one. This would be expected for points in different crowns of varying height, whereas points at the top and bottom of the same crown would expect to have vertical components that point towards a central point in the crown. As with the horizontal comparison, this reduction is larger for points which have a smaller vertical distance, $\mathrm{d}_{\mathrm{Z}}$, between them. The normalisation for this, $K_{\mathrm{Z}}$, is set to half of the aboveground height of the tallest point of the data. The reduction is also larger when the relative difference in the vertical centroid vectors is larger. An example of this reduction is shown in Figure \ref{fig:geom_ex}f. With this, the final similarity weights, $w_{ij}$, as shown in Figure \ref{fig:geom_ex}c, are computed as

\begin{equation}
\begin{aligned}
w_{ij} = &
\begin{cases}
w_{ij}^{\mathrm{Z}}, & \text{if } \mathrm{sign}(\Delta_{i}^\mathrm{Z}) \neq \mathrm{sign}(\Delta_{j}^\mathrm{Z}) \text{ and taller point has $\Delta^{\mathrm{Z}}\geq0$}, \\
w_{ij}^{\mathrm{postH}},&\text{otherwise}.
\end{cases}
\\
\text{with } & w_{ij}^{\mathrm{Z}} = \hspace{0.5em} w_{ij}^{\mathrm{postH}} \times \mathrm{exp}\bigg(-W_{\mathrm{Z}} \times \frac{K_{\mathrm{Z}}}{\mathrm{d}_{\mathrm{Z}}}\times|\Delta_{i}^\mathrm{Z}-\Delta_{j}^\mathrm{Z}|\bigg).
\label{eq:w_Z}
\end{aligned}
\end{equation}

Calculating pairwise weights for every set of two points would be computationally cumbersome, producing a matrix of weights that is far too large to store in a typical computer\textsc{\char13}s memory ($\leq16$ GB RAM). As an indication, a point cloud of 40,000 points would require 12.8 GB of RAM to be held in memory, ignoring overheads and the need for spare memory to perform the necessary computations, which at the point density of the Sepilok dataset would cover roughly 0.5 ha. To resolve this issue, the Nystr\"om extension is used \cite{Fowlkes2004}. Here the complete matrix and relevant eigenvectors are computed for a subset of the points, typically less than 10\% of the data. The eigenvectors are then extended to the full dataset through quadrature based on the weights in a manner that produces a robust approximation of $w_{ij}$\cite{Fowlkes2004,Bertozzi2012}. This allows computation of all pairwise weights for the subset and avoids a need for a sparse representation of the pairwise linkages.

\subsubsection{Using crown allometry to refine the segmentation }
\label{sssec:meth_tcs_post}
An extension of the basic graph cut algorithm makes use of knowledge of crown radius scaling with tree height to remove improbable trees, by post-processing of clusters identified by the initial graph cut. For a candidate tree in the segmented dataset, a `maximum' predicted crown radius was taken from look-up table based on its height. The numbers in the look-up table were based on the 95th percentile relationship form the global database in [\citen{Jucker2017}] as given in equation \eqref{eq:allom_H_CD_TF_me}. Any crowns which exceeded the range of this dataset were set to match the maximum allometric radius for the dataset. First candidate crowns that overlapped too much with larger neighbours, both in horizontal extent and vertical overlap were merged. Excessive horizontal overlap occurs when the lower crown has its tree top within the taller crown's allometric radius or when 60\% or more of the points in the lower crown are within the radius of the upper crown. Excessive vertical overlap occurs when the bottom 25\% of points in the upper crown are below the top 25\% of points in the lower crown. Crowns are only merged if both horizontal and vertical overlap are excessive. Then crowns where more than 5\% of points lie beyond the computed maximum radius from the tree top are further trimmer. Trimming uses hierarchical clustering based on the Euclidean distance between points, using raw heights. Two clusters are produced and the one which includes the tree top is kept, with points in the other cluster added to the rejected points list. Finally crowns which contain too few points were rejected, with a threshold of a minimum of 100 points per crown. This was set to filter crowns which were missing many points and to avoid the trimming step from causing crowns to be too small. The goal here is to ensure only allometrically feasible crowns are kept, with all other points being marked as not contained within a crown. These points are rejected in this application of graph cut.

\subsubsection{Detecting lower-canopy trees}

\begin{figure}[!t]
\begin{center}
\includegraphics[width=1\textwidth]{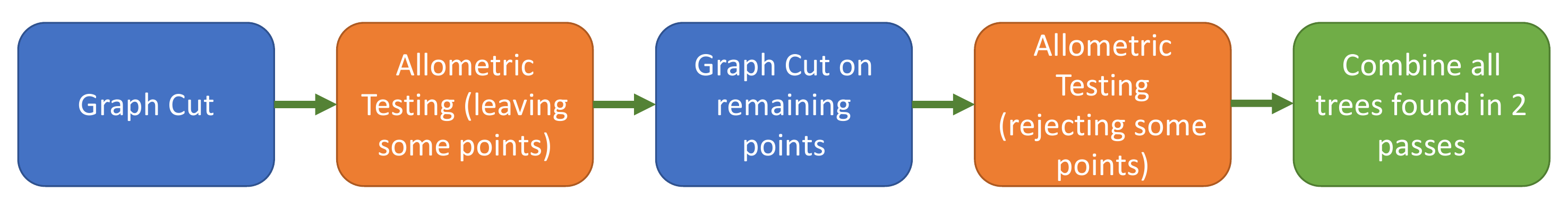}
\caption{Pipeline for double-layer approach to MCGC}
\label{fig:dbl_diagram}
\end{center}
\end{figure}

The allometric refinement stage leads to there being points which are not assigned to any tree. These can be from trees which are close neighbours of successfully detected crowns where removal of this crown makes delineation easier. Equally many of these points are in the understory, where point densities are lower due to occlusion and differentiation of crowns is more challenging. A second pass of MCGC is then used to detect tree crowns from the unassigned points. These trees are then added to those accepted in the first application of MCGC to produce the final list of crowns. It is possible to alter the weight parameters to reflect differences in the canopy structure in these lower layers; however we applied the MCGC algorithm with the same set of parameters in the second pass in this study. This simple extension is illustrated in Figure \ref{fig:dbl_diagram}.

\subsubsection{Selecting parameters}
\label{ssec:meth_param}

The values of $\sigma_{\mathrm{XY}}$, $\sigma_{\mathrm{Z}}$, $W_{\mathrm{H}}$ and $W_{\mathrm{Z}}$ in equations \eqref{eq:base_w} -- \eqref{eq:w_Z} control the relative importance assigned to each component in computing similarities between points. To set these, nine of the 36 1-ha plots were used, three from each soil type, comprising 25\% of the data in this study. To decide the best choice of parameters from those trialled, the resulting segmentations were manually inspected, and the distribution of detected crown sizes was qualitatively compared to the distribution of field inventory stem sizes for the 9 plots. First the values of $\sigma_{\mathrm{XY}}$ and $\sigma_{z}$ were simultaneously assessed, by trying MCGC with each parameter taking one of a number of values and trying all possible pairings. For this process, the modifications in equations \eqref{eq:w_H} and \eqref{eq:w_Z} were not applied to any weightings. It was expected that the vertical distance would need a smaller parameter, penalising this more heavily, as tropical crowns are often wide and less vertically extended than the typical conic shape of coniferous forests\cite{Jucker2017}. Analogous parameters in previous work on the use of graph cut on coniferous forests found the horizontal parameter to be the smaller, following the same reasoning\cite{Lee2017}. Similarly to the first two parameters, $W_{\mathrm{H}}$ and $W_{\mathrm{Z}}$ were simultaneously trialled in an exhaustive manner, with $\sigma_{\mathrm{XY}}$ and $\sigma_{\mathrm{Z}}$ set to their selected values. The final parameters used in this study are listed in Table \ref{table:param}.

\begin{table}[tp]
\caption{Values of parameters selected for MCGC by trial and error}
\begin{center}
\begin{tabular}{ccccc}
\toprule
& $\sigma_{\mathrm{XY}}$ & $\sigma_{\mathrm{Z}}$ & $W_{\mathrm{H}}$ & $W_{\mathrm{Z}}$\\
\cmidrule{2-5}
Value of parameter & 4 & 2 & 0.2 & 0.2\\
\bottomrule
\end{tabular}
\end{center}
\label{table:param}
\end{table}

\subsection{Improving computational efficiency}
\label{subsec:meth_eff}

As shown in Section \ref{sec:frame}, applying the graph cut algorithm amounts to solving an eigenvector problem. Such problems do not scale well with increasing matrix size. The Nystr\"om extension already reduces the effective matrix size used in the algorithm. However, as the number of points in the dataset increases, the total memory required still increases. To resolve this only with the Nystr\"om exension would require taking a very small subset of points for applying the approximation. Instead, when working with datasets of 1 ha or more a simple method for reducing the workload was developed, by downsampling the data and then imputing the full data results. This can be justified as the dense point cloud is locally correlated -- points close in horizontal extent will have similar height values and a subsampled point cloud retains the 3D structure of the dense data. This way the MCGC algorithm is applied in full to a subsampled point cloud and the Nystr\"om extension can be applied at the same proportion of points used in MCGC for datasets over large areas, or from very dense ALS data.  The full pre-processed point cloud is downsampled, by random (without replacement) sampling of the data. The Multi-Class Graph Cut is then applied to this subset. Once the crowns are identified, the full dataset is then imputed by a $m$-nearest neighbour approach. The value of $m$ is set to match the effective downsampling, so if 1/5 of the data is used then $m=5$ is used for imputation. For a given crown, points could only be added to it in imputation if they lay within the maximum allometric radius of this crown, as computed from the 95th percentile relationship in equation \eqref{eq:allom_H_CD_TF_me} for the height of this crown. To ensure imputation did not create artefacts, where rejected crowns were now grouped with their nearest valid crown, the DBSCAN algorithm was applied to each final tree\cite{Ester1996}. In our approach this finds groups of points for which there are at least 10 other points within 2 m of the considered point. A connected neighbourhood is then constructed as all neighbouring points which satsify this property, as well as all points within 2 m of any point in this group. This ensured imputed crowns are formed of a single locally-connected group of points and any points not connected to this were marked as unassigned. In this work we used a subsampling pool of 20\% of the data. In the double-layer extension, subsampling and imputation was applied to each application of MCGC, meaning it was applied twice in the complete pipeline. In running the full double-layer MCGC algorithm to all 36 one ha plots in this study, the total time taken averaged 66,250 s. This is equivalent to the algorithm taking 30 m 40 s per plot. This timing was completed on a workstation running Windows 7 using MATLAB 2017a. The workstation was equipped with an Intel Xeon E3-1240 V2 CPU, comprising 8 cores running at 3.4 GHz with 16 GB RAM. MATLAB allocates memory smartly to enable calculations to proceed where theoretically they may be RAM-limited. Accordingly, running the MCGC algorithm on a machine with more RAM, or where subsampling is optimised to adapt to the RAM restrictions of a machine has the potential to accelerate the algorithm, but we do not explore that option in this work.

\subsection{Assessment of segmentation accuracy}
\label{subsec:meth_acc}
 
MCGC was applied to the 36 1 ha plots within the Sepilok dataset. The two versions outlined in Section \ref{subsec:meth_tcs} of the algorithm were tested: (1) \textbf{single-layer MCGC} applied a single graph cut to the data followed by the allometric filtering (as per Figure \ref{fig:gc_fullalg}); (2) \textbf{double-layer MCGC} extended the results of Simple MCGC by applying a second pass of the algorithm to unassigned points (as per Figure \ref{fig:geom_diagram}).

To compare the distribution of trees found by the automatic detection methods, the diameter at breast height (DBH) of these were estimated. This metric was previously recorded in the Sepilok data set for all stems, whereas their heights were scarcely recorded. To convert from remotely sensed height to DBH, log-log regression was applied to the 91 trees manually delineated by Coomes et al.\cite{Coomes2017}. This results in the following relationship for $\mathrm{H}$ (m) and $\mathrm{DBH}$ (cm):
\begin{equation}
\mathrm{DBH} = 0.252 \times \mathrm{H}^{1.465}
\label{eq:allom_D_H_91}
\end{equation}
DBH for each tree crown detected by the MCGC algorithms was estimated from their above-ground height using this relationship. The results were then compared by grouping stems into a number of diameter classes and comparing totals.

\subsection{Predicting biomass}
\label{ssec:meth_biomass}

Field surveys are commonly used as a basis for estimating carbon stored in above-ground biomass. Accordingly we computed biomass estimates for the trees in each plot to measure the total estimated above-ground carbon density for each 1 ha plot (ACD, $\text{Mg}\hspace{0.1em}ha^{-1}$). For each plot we compared the total ACD from the field survey to that from each of the MCGC based methods. Here AGB was computed for each tree, and these values were then summed for each plot. Finally a conversion factor of 0.47 was applied to convert from AGB to ACD\cite{Martin2011}.

For the field survey, the biomass from each tree in the forest inventory was computed according to Chave et al.'s pantropical equation\cite{Chave2014}:

\begin{equation}
\label{eq:AGB_field}
\mathrm{AGB}_{\mathrm{field}} = 0.0673 \times (\mathrm{WD} \times \mathrm{D}^{2} \times \mathrm{H})^{0.976},
\end{equation}

\noindent
where $\mathrm{WD}$ is the wood density as taken from the global wood density database\cite{Chave2009,Zanne2009}, $\mathrm{D}$ is the stem diameter (cm) and $\mathrm{H}$ is the estimated height (m) based on equation \eqref{eq:allom_H_D}. Wood density was mapped to the best taxonomic unit available for each species. If there was not data for a species, then the average for its genus was used, and similarly where there was no data for this the family average was used. If there was no data at family level, an average of all species present in the Sepilok plots was used. The AGB per plot was then computed by summing the contribution from each tree before being converted to ACD. 

The ACD contribution of each automatically segmented tree was computed using the following relationship, originally derived by Coomes et al. in [\citen{Coomes2017}] from 91 crowns manually delineated and verified in the LiDAR data:
\begin{equation}
\label{eq:AGB_auto}
\mathrm{ACD}_{\mathrm{auto}} = 0.268 \times ( \mathrm{H} \times \mathrm{CD})^{1.45},
\end{equation}
\noindent
where $\mathrm{H}$ is the aboveground height of the segmented tree (m) and $\mathrm{CD}$ is the crown diameter (m) computed from the crown area ($\mathrm{CA}$) as $\mathrm{CD} = 2\times\sqrt{\mathrm{CA}/\pi}$. For the MCGC output, the crown area was taken to be the area of the convex polygon enclosing each tree. Comparisons between the ACD estimates from the field inventory and remote sensing methods were compared for each one ha plot. Overall results are then reported via the bias and Root Mean Square Error (RMSE) of the predictions. These were computed as a percentage of the predictions of ACD for the field inventory data. Here a negative bias indicates the remote sensing estimate is an underestimate.

\subsection{Comparison to existing methods}
\label{ssec:meth_comaparison}

MCGC was compared to two existing algorithms as a benchmark of performance. The itcSegment algorithm as implemented in [\citen{Coomes2017}] was used as a reference ITC algorithm. This is applied to the rasterised CHM, computed at a 0.5 m resolution. First a moving window local-maxima finder detects tree tops, where the size of the window scales with above ground height in the same manner as our prior generation step does. Finally trees are delineated by a region-growing algorithm taking the local maxima as seed points. Data from this algorithm were processed in the same way as those from MCGC as outlined in Sections \ref{subsec:meth_acc} and \ref{ssec:meth_biomass}. Estimation of AGB was further compared to an area-based model\cite{Asner2014}. This predicts biomass at one ha scale based on metrics of the CHM. The approach uses the general biomass equation from [\citen{Asner2014}] for which ACD is computed for a one ha plot as $\mathrm{ACD_{General}} = 3.836 \times \mathrm{TCH}^{0.281} \times \mathrm{BA}^{0.972} \times \mathrm{WD}^{1.376}$, where TCH is mean top canopy height (m), BA is basal area ($\mathrm{m}^2 \hspace{0.2em} \mathrm{ha}^{-1}$) and WD is diameter-weighted mean wood density ($\mathrm{g \hspace{0.2em} cm}^{-3})$. To estimate BA and TCH from the ALS data the approach used in [\citen{Coomes2017}] was applied, where both  were estimated as power laws of TCH based on the 36 one ha plots to give an overall model for ACD as $\mathrm{ACD_{General}} = 7.37 \times \mathrm{TCH}^{0.870}$. Following the suggestions in [\citen{Asner2014}] and replicating the approach in [\citen{Coomes2017}] a locally-fitted area-based model was also used. This takes the same form as the general model but with the powers and multiplicative factor directly fitted to the Sepilok plots but log-log regression. Further, following the work in [\citen{Coomes2017}], BA was predicted not from TCH, but instead from Gap Fraction at a height of 19 m ($\mathrm{GF}_{19}$). This counts the number of pixels in the CHM raster for which the TCH is below 19m as a percentage of all pixels in the raster. Brought together these produce a locally-fitted model for ACD as $\mathrm{ACD_{Local}} = 25.93 \times \mathrm{TCH}^{0.437} \times \mathrm{GF}_{19}^{-0.209}$.

\section{Experimental Results}
\label{sec:results}

\begin{figure}[!t]
  \begin{center}
  \includegraphics[width=\textwidth]{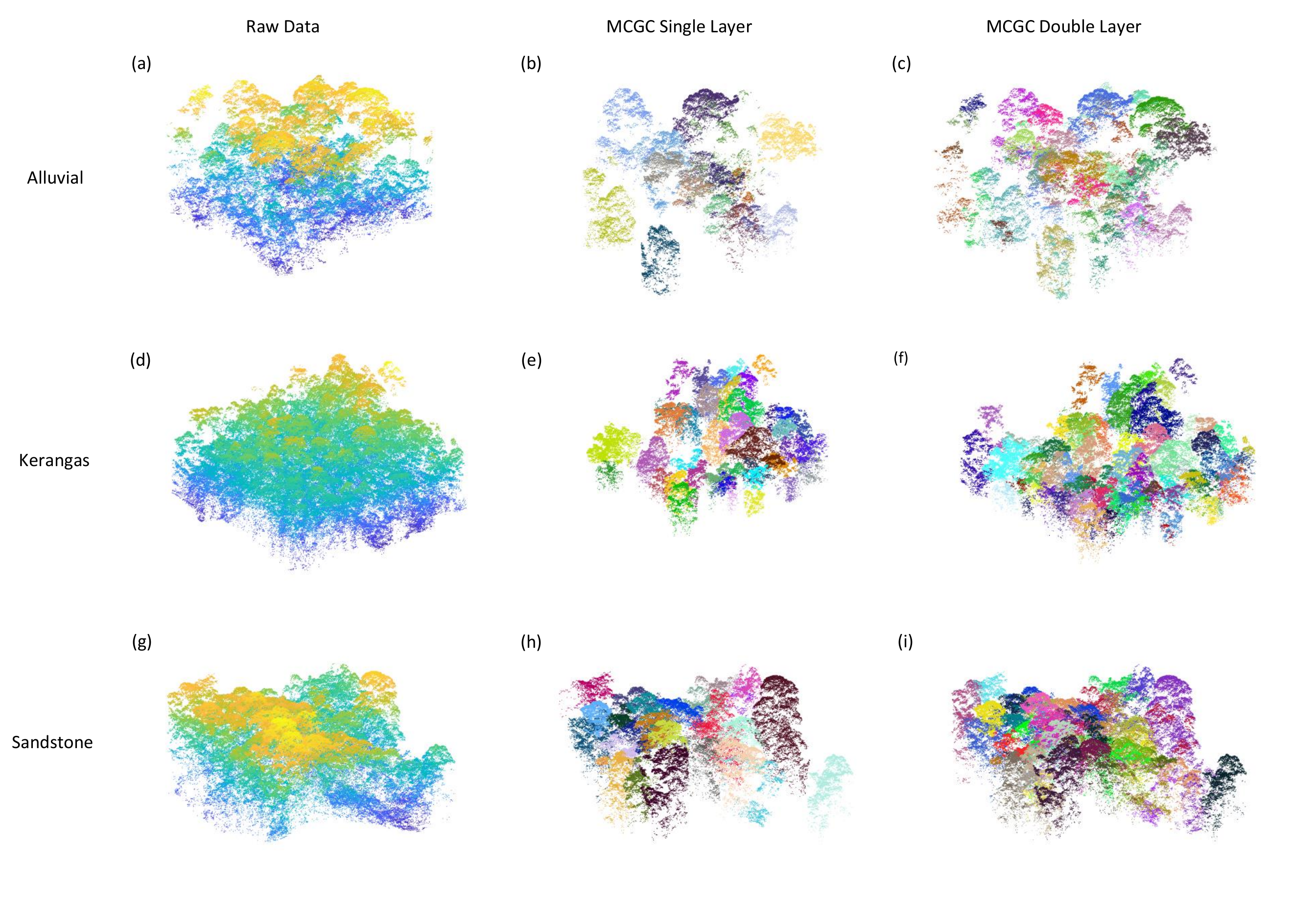}
       \caption{Example results of applying MCGC single-layer and MCGC double-layer to a 1 ha plot of each forest type. Each row represents a plot of one of the three soil types: Alluvial (a--c), Kerangas (d--f) and Sandstone (g--i). Across each row are the raw data (coloured by height), the output from single-layer MCGC (coloured by crown) and the output from double-layer MCGC (coloured by crown). Points which are unassigned are not included in the MCGC output images and the increased detection of trees can be seen by comparison between the second and third columns: pairs b \& c, e \& f and h \& i.}
    \label{fig:res_pic}
	\end{center}
\end{figure}

\subsection{Tree detection}
\label{subsec:res_tree}

\begin{figure}[!t]
  \begin{center}
      \includegraphics[height=0.8\textheight]{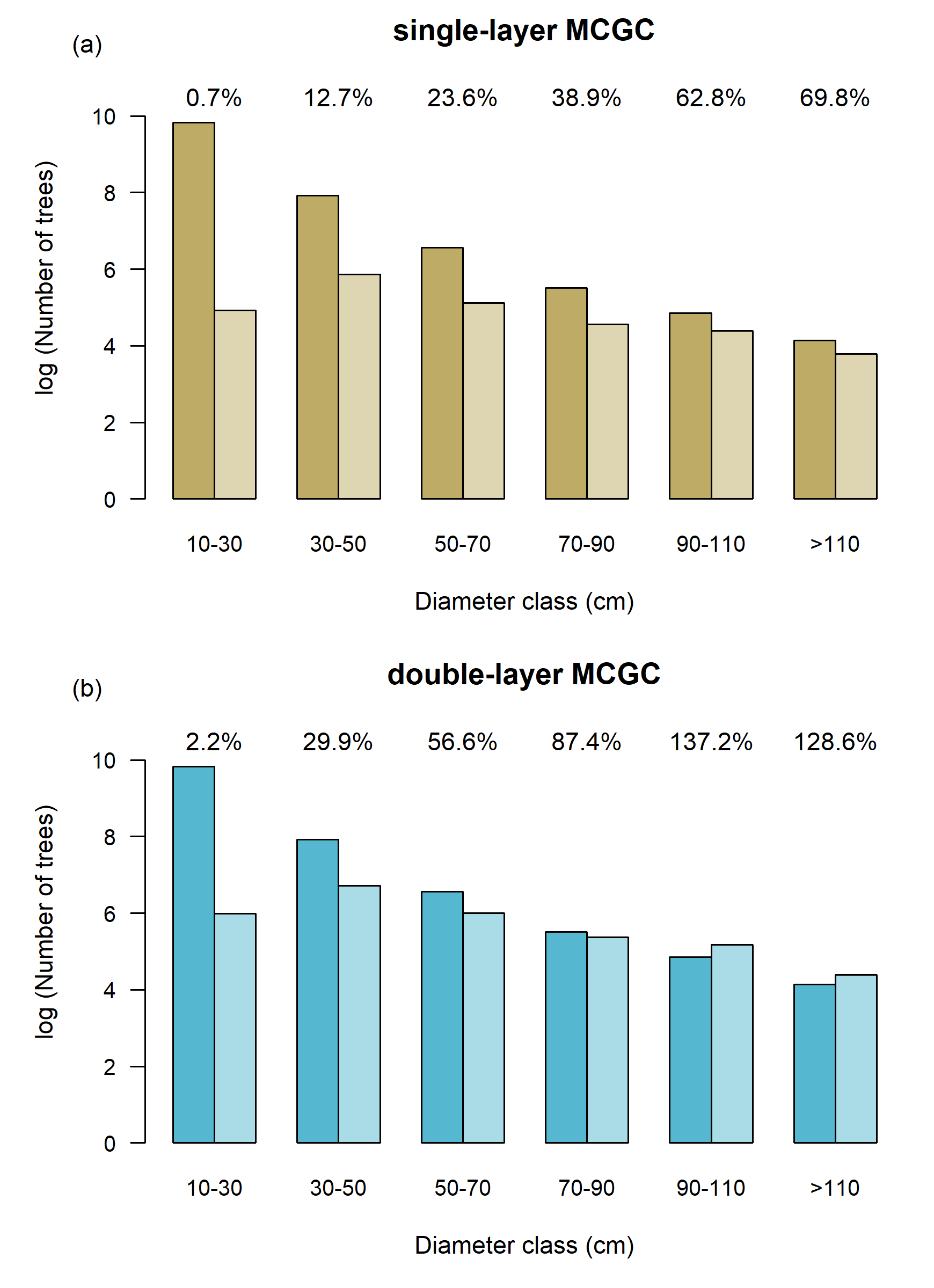}
      \caption{Comparison of log-transformed numbers of trees by stem diameter for field inventory, left and dark, and each MCGC implementation, right and light. Percentages are given for MCGC as fraction of field inventory. MCGC methods are: (a) single-layer MCGC, (b) double-layer MCGC. Applying the second round of MCGC increases detection rates across all stem size groups and more than doubles the rate of detection in stems of a size of 90 cm or less.}
       \label{fig:res_DvD}
	\end{center}
\end{figure}

 Single-layer MCGC detects many of the largest trees, but misses intermediate and lower layers of the canopy.  The additional detection of trees in double-layer MCGC is evident in the example results for one plot of each forest type shown for each method in Figure \ref{fig:res_pic}.  Figure \ref{fig:res_DvD} shows the comparison between the field survey measurements and MCGC-estimated DBH values broken down by bands of diameter and detection rates by size class are summarised and compared with the results from using itcSegment in [\citen{Coomes2017}] in Table \ref{table:DD}. Single-layer MCGC under-estimated the number of trees across all size classes (Figure \ref{fig:res_DvD}a). This results from a strict set of allometric testing criteria, where clusters that don't pass are rejected. Thus under-estimation of even the tallest trees can be expected. In this case, many of the tallest trees are accounted for (69.8\% of trees with DBH$>$ 110 cm and 62.8\% of those with 90 cm$<$ DBH $<$110 cm). However, when compared to the same counts for itcSegment, which works only on the top canopy surface, (103.2\% and 82.2\% respectively) it is clear single-layer MCGC does not account for all trees in the top layer as these are being detected by itcSegment. For all trees with DBH of 90 cm or smaller a single application of MCGC detects fewer than half the number of stems recorded in field inventory, and itcSegment outperforms single-layer MCGC. These trees account for the vast majority of total stems though many will be in the intermediate layers of the canopy. This motivates and justifies applying the algorithm a second time to detect these intermediate layer trees with many of the largest trees already confidently detected. 
 
Once a second pass of MCGC was applied in the double-layer MCGC pipeline (Figure \ref{fig:res_DvD}b) the number of trees found increased across all size classes. This leads to some overestimation in the tallest classes (128.6\% for DBH $>$ 110cm and 137.2\% for 90 cm$<$ DBH $<$110 cm) but leads to much better estimates of trees in the intermediate size ranges (87.4\% for 70 cm$<$ DBH $<$90 cm and 56.6\% for 50 cm$<$ DBH $<$70 cm). These detection rates are comparable to, or improvements on, itcSegment for the same size stems (51.8\% and 61.9\% respectively). Double-layer MCGC still misses most of the trees in the lowest canopy layers (29.9\% for 30 cm $<$ DBH $<$ 50 cm and 2.2\% for 10 cm $<$ DBH $<$ 30 cm). In these classes of stem size itcSegment reports a higher number of stems though itself still misses a large proportion of stems (53.5\% and 9.5\% respectively). For all size classes where DBH $<$ 90 cm, adding a second pass of MCGC more than doubles the counts of stems found by single-layer MCGC which means more trees in these size classes are found in the second application of MCGC than are found in single-layer MCGC alone.

\begin{table}[tp]
\caption{Summary of detection rates by size class for MCGC and itcSegment}
\begin{center}
\begin{tabular}{ccccccc}
\toprule
 & \multicolumn{6}{c}{Detection Rate (\%)}  \\
 & \multicolumn{6}{c}{Size class (cm)}  \\
Algorithm & $>$110 & 90-110 & 70-90 & 50-70 & 30-50 & 10-30\\
\cmidrule{1-7}
itcSegment\cite{Coomes2017} & 103.2 & 82.2 & 51.8 & 61.9 & 53.5 & 9.5  \\
single-layer MCGC & 69.8 & 62.8 & 38.9 & 23.6 & 12.7 & 0.7 \\
double-layer MCGC & 128.6 & 137.2 & 87.4 & 56.6 & 29.9 & 2.2 \\
\bottomrule
\end{tabular}
\end{center}
\label{table:DD}
\end{table}

\subsection{Biomass estimation}
\label{subsec:res_biomass}

\begin{figure}[!t]
  \centering
  \includegraphics[width=\textwidth]{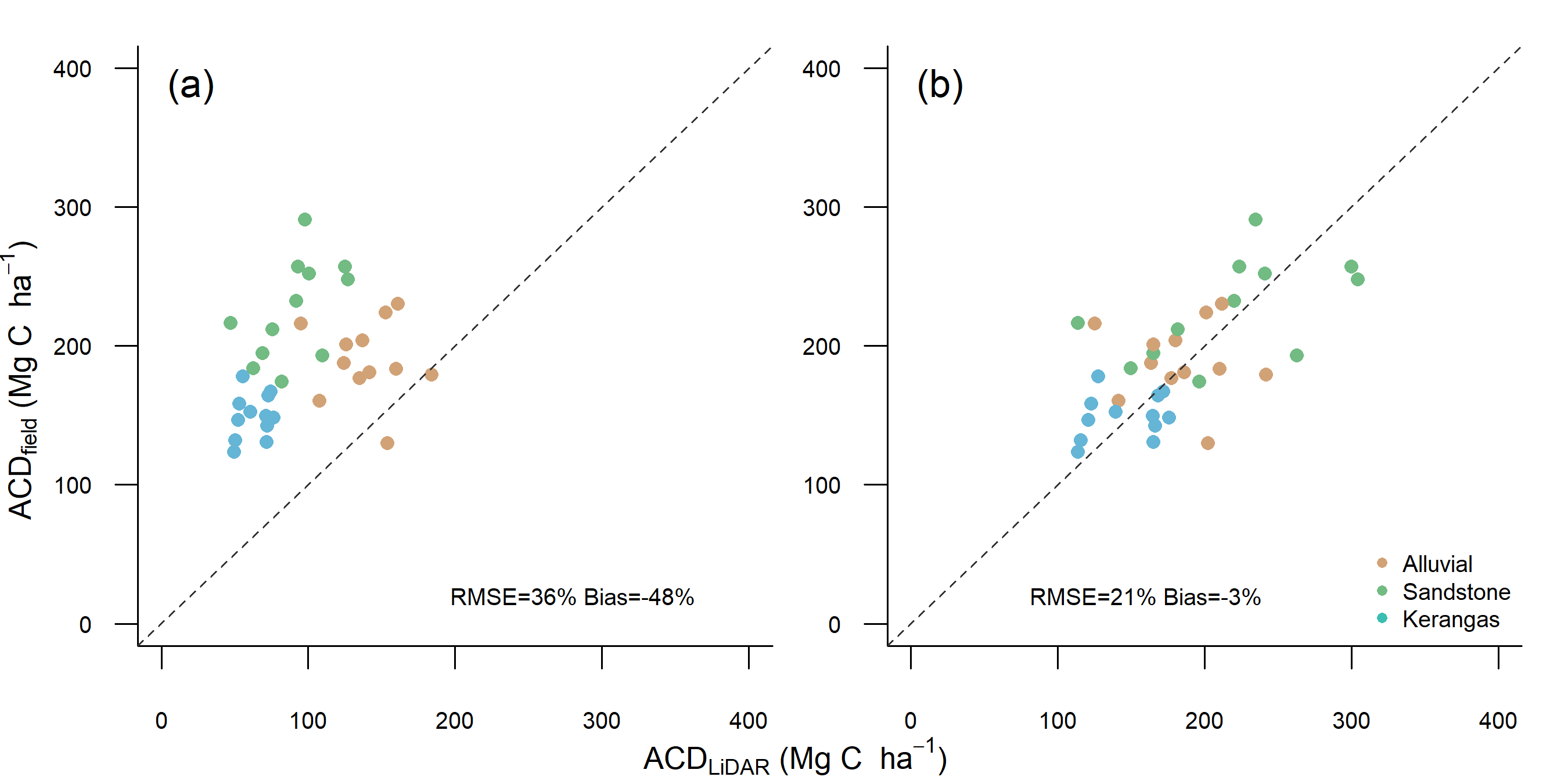}
  \caption{Comparison of field-estimated above-ground carbon density (ACD\textsubscript{field}) with estimates based on single-layer Multi-Class Graph Cut (ACD\textsubscript{LiDAR}) for each of the 36 one-ha subplots. Panel (a) shows the original estimates with a 1:1 line (dashed). (b) shows the effect of applying a linear forest type specific correction factor to ACD\textsubscript{LiDAR}.}
   \label{fig:res_ACD_sgl}   
\end{figure}

\begin{figure}[!t]
  \begin{center}
  \includegraphics[width=\textwidth]{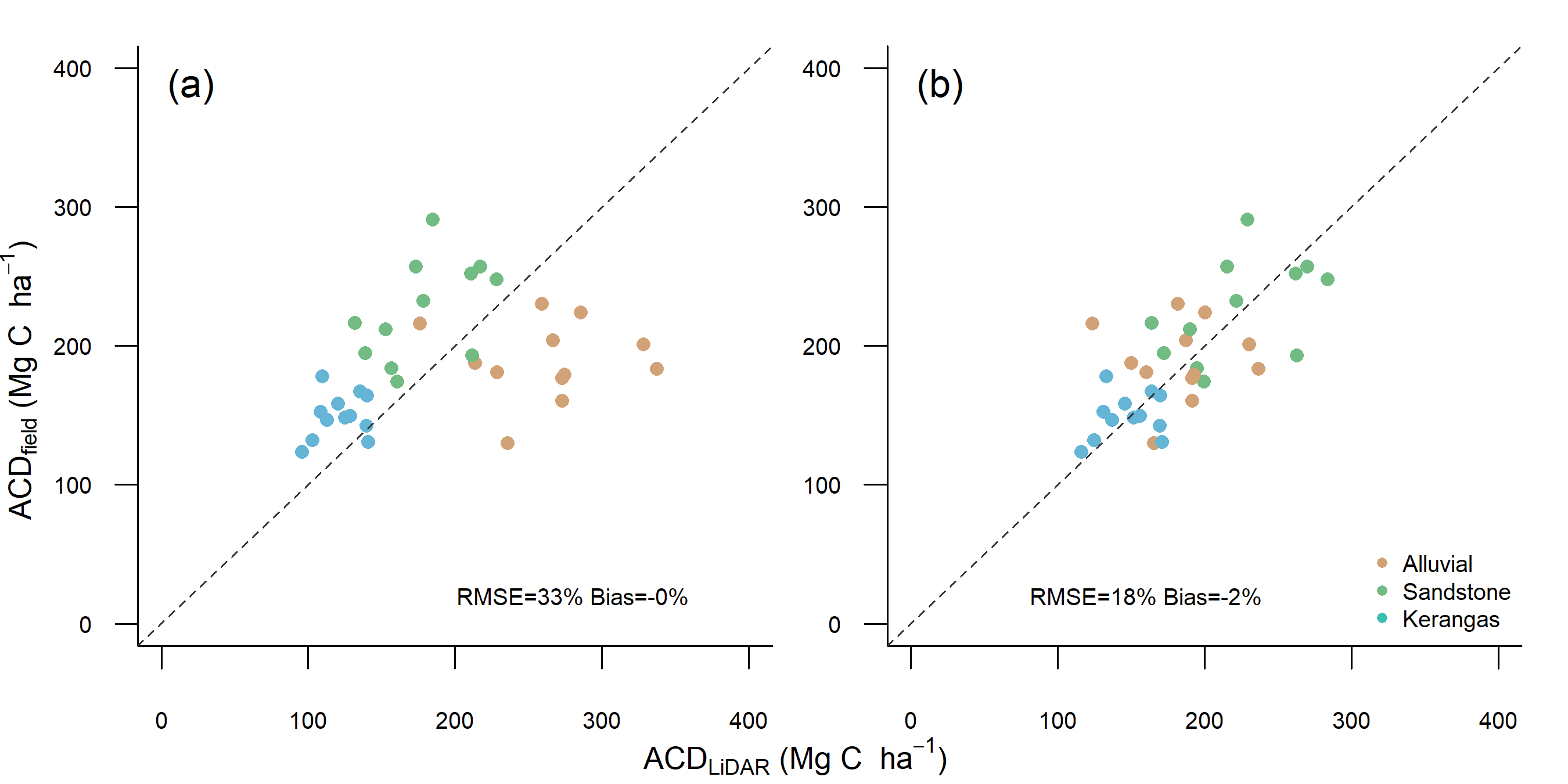}
       \caption{Comparison of ACD estimates by plot for double-layer MCGC for each of the 36 one-ha subplots. ACD\textsubscript{field} and ACD\textsubscript{LiDAR} are the ACD estimates from the field survey and from MCGC applied to LiDAR data respectively. (a) shows the original estimates with a 1:1 line (dashed). (b) shows the effect of applying a lieane forest type specific correction factor to ACD\textsubscript{LiDAR}.}
    \label{fig:res_ACD_dbl}
	\end{center}
\end{figure}

\begin{table}[tp]
\caption{Bias and RMSE for AGB estimates from MCGC, itcSegment and Area-based Modelling for both original model output and once a forest type specific correction factor has been applied to MCGC and ITC and the area-based model has been locally calibrated}
\begin{center}
\begin{tabular}{ccccc}
\toprule
& \multicolumn{2}{c}{Original} & \multicolumn{2}{c}{Specifically Calibrated} \\
Algorithm & Bias (\%) & RMSE (\%) & Bias (\%) & RMSE (\%) \\
\cmidrule{1-5}
itcSegment\cite{Coomes2017} & -20 & 26 & -1 & 18 \\
Area-based Modelling\cite{Coomes2017} & -19 & 20 & 0 & 13 \\
single-layer MCGC & -48 & 36 & 3 & 21 \\
double-layer MCGC & -0 & 33 & -2 & 18 \\
\bottomrule
\end{tabular}
\end{center}
\label{table:AGB}
\end{table}

\begin{table}[tp]
\caption{Bias and RMSE for AGB estimates from MCGC, itcSegment and Area-based Modelling based on original model output broken down by forest type}
\begin{center}
\begin{tabular}{ccccccc}
\toprule
 & \multicolumn{2}{c}{Alluvial Forest} & \multicolumn{2}{c}{Kerangas Forest} & \multicolumn{2}{c}{Sandstone Forest}\\
Algorithm & Bias (\%) & RMSE (\%) & Bias (\%) & RMSE (\%) & Bias (\%) & RMSE (\%)\\
\cmidrule{1-7}
itcSegment\cite{Coomes2017} & -14 & 20 & -19 & 9 & -29 & 18 \\
Area-based Modelling\cite{Coomes2017} & -3 & 15 & -21 & 8 & -30 & 13 \\
single-layer MCGC & -26 & 23 & -58 & 17 & -60 & 22 \\
double-layer MCGC & 39 & 22 & -18 & 14 & -21 & 16 \\
\bottomrule
\end{tabular}
\end{center}
\label{table:AGB_for}
\end{table}

Single-layer MCGC underestimated biomass of all three forest types (bias: -48\%, Figure \ref{fig:res_ACD_sgl}a and Table \ref{table:AGB}) and including a second pass of MCGC removed the overall bias in estimating plot ACD Figure \ref{fig:res_ACD_dbl}a and Table \ref{table:AGB}). This is consistent with the comparison of stem counts in Figure \ref{fig:res_DvD}a which showed that the single-layer MCGC approach underestimated stem sizes across all classes. Including a second pass of the MCGC algorithm removed the overall bias in the predictions (Figure \ref{fig:res_ACD_dbl}a). However, this resulted from a tendency to overestimate biomass in the Alluvial forest type matched with small underestimation in both Kerangas and Sandstone forest types (Table \ref{table:AGB_for}). Underestimation can be accounted for by missing trees in the lower canopy. Alluvial forest contains almost all of the tallest trees in Sepilok and so the tendency of double-layer MCGC to over-detect these explains the overestimation of ACD in these plots. Both approaches had a relatively large RMSE (36\% and 33\%) but this is not unexpected in an individual tree approach to estimation of biomass as big trees contain a large proportion of the total biomass. Small errors in prediction of the largest trees can therefore have a large effect on overall predictions of AGB and ACD. By comparison, in the original study by Coomes et al. in [\citen{Coomes2017}], the general area-based approach for estimating ACD, as per [\citen{Asner2014}], produced predictions with a bias of -19\% and RMSE of 20\% (Table \ref{table:AGB}). The ITC approach in this study produced ACD predictions with a bias of -20\% and RMSE of 26\% (Table \ref{table:AGB}). When looking at the contributions of individual forest types, the methods in the original study produce closer estimates for Alluvial forest types, as a result of the overestimation by double-layer MCGC (Table \ref{table:AGB_for}). In contrast, for both Kerangas and Sandstone forest, double-layer MCGC produces closer estimates to the field inventory than any of the other methods considered here.

Previous work on individual-based AGB estimation notes that it is common for such approaches to systematically under-estimate total biomass as trees in the lowest layers of the canopy are rarely detected\cite{Dalponte2016,Coomes2017}. It can therefore be informative to apply a simple linear correction factor to account for this. In [\citen{Coomes2017}] this was applied both to all plots simultaneously and with a different correction factor applied to each forest type. As we found our overall estimate of ACD from double-layer MCGC was already low-bias we chose to only investigate the effect of a forest-type specific correction factor. The MCGC algorithms here used only a regional allometry which was based on data for the whole Indo-Malaya region (equation \eqref{eq:allom_H_CD_TF_me}). There was no local allometric input for the MCGC algorithm nor any forest-type specific step. It is therefore informative to apply this correction step. A forest-type specific correction was computed by fitting a linear regression model to the ACD estimates from each MCGC implementation and the field inventory for the 12 one-hectare plots of each forest type, constrained to pass through the origin. For single-layer MCGC the correction factors were larger than 1 as expected given the algorithm underestimated ACD (Alluvial 1.32, Sandstone 2.40, Kerangas 2.30). Double-layer MCGC had correction factors that showed a small underestimation of ACD in Kerangas and Sandstone forest matched with a similar overestimation in alluvial forests (Alluvial 0.70, Sandstone 1.24, Kerangas 1.21). Applying this correction removed the bias for single-layer MCGC (3\% vs 48\%), but only reduced the bias within each forest type for double-layer MCGC as this already had low overall bias  (Tables \ref{table:AGB}). In both cases this reduced the RMSE as any bias for each forest type was reduced independently (21\% vs 36\% and 18\% vs 33\% for single-layer and double-layer MCGC). The effect of these corrections are similar for itcSegment, removing bias and reducing RMSE, as reported in [\citen{Coomes2017}] and shown in Table \ref{table:AGB}. When choosing to use a locally-calibrated model for the area-based approach the same effects are seen again. Here the model has all parameters directly fitted to the data at Sepilok, as well as choosing the Gap Fraction height that best predicts wood density as per the work in [\citen{Coomes2017}].

\section{Discussion}
\label{sec:discussion}

\subsection{Assessment of MCGC Performance}
\label{subsec:dis_MCGC}

\subsubsection{Forest inventory}
\label{ssubsec:dis_MCGC_FI}

When comparing the distribution of crown sizes found by single-layer MCGC to the reference field inventory it is clear that across all stem sizes the algorithm did not find all individual trees (Figure \ref{fig:res_DvD}a). The initial graph cut segmentation is constrained to find at least as many crowns as a simple maxima finding algorithm. As shown in Table \ref{table:DD}, the itcSegment algorithm, which starts with the same local maxima approach finds more than 50\% of the count of stems for trees with DBH $>$ 30 cm. These totals represent the number of local maxima initially found as itcSegment is constrained to delineate one tree for each local maximum. From this single-layer MCGC then found fewer stems than itcSegment. The graph cut step is constrained to find at least as many trees as the initial prior number of trees. Therefore the underestimation likely stems from the allometric filtering step. This step applies multiple criteria to each crown, based on knowledge of a regional allometry, rejecting those crowns that do not meet the criteria closely enough. This results in a reduction of the number of crowns, but those that remain are allometrically feasible. Notably, the algorithm finds a greater proportion of crowns relative to itcSegment in the largest classes (DBH $>$ 70 cm) than the smaller classes, suggesting that the rejection of crowns may be more common for medium and small trees. 

Double-layer MCGC detects more tree crowns in all size classes (Figure \ref{fig:res_DvD}b and Table \ref{table:DD}). These additional crowns are detected in the second application of the MCGC algorithm. The algorithm over-detects the number of trees in the largest size classes (DBH $>$ 90 cm), though there are fewer trees of these sizes, meaning the over count is not a large number of trees compared to the total count of all stems. As a trade-off, double-layer MCGC is able to detect more than double the number of trees in all small and medium classes (DBH $<$ 90 cm) when compared to the single-layer approach. This reduces the under counting in these size classes. This results from the `stripping-off' of the dominant trees to allow the algorithm to better segment the under layers of the canopy. However the algorithm struggles with finding the smallest trees as is a noted drawback of working with ALS data\cite{Coomes2017}. The number of returns for these trees is very small and this underlines the importance of alternative approaches, such as using TLS data, for mapping the lowest layers of the canopy. Overall double-layer MCGC is able to detect most crowns for intermediate and large sized trees.

Results from the itcSegment approach in [\citen{Coomes2017}] justify the choice of limits for the choice of total tree stems. These limits are set to a minimum of the number of local maxima itcSegment would find and a maximum of twice this. From Table \ref{table:DD} it is clear that itcSegment underestimates the number of crowns in all but the largest stem class. Therefore the total number calculated in this way will be an underestimate, though estimates well for trees of DBH $>$ 90 cm. Equally, for all stems where DBH $>$ 30 cm, itcSegment finds at least 50\% of the total number of stems for each size class. Therefore doubling the number of local maxima found by the first step of this algorithm should over estimate the number of stems in all size classes where DBH $>$ 30 cm. This would then make a sensible upper limit for the number of crowns to seek. For trees where DBH $<$ 30 cm all such approaches result in under-counting. These trees are mostly found in the lower layers of the canopy and, as discussed below, this is a persistent problem with ALS data.

The difficulty of finding trees in the lowest layers of the forest canopy is a persistent problem in analysing ALS data in multi-layered canopies \cite{Vauhkonen2012,Hamraz2017,Coomes2017}. Here occlusion causes a diminishing number of returns for lower canopy, making it hard to identify the smallest trees in the subcanopy, even when inspecting the data visually\cite{Vauhkonen2012,Hamraz2017}. This highlights the importance of taking a combined approach to data collection, pairing ALS data acquisition with other methods such as TLS or field inventory. These methods themselves suffer from limitation in scope of the area they can cover when compared to ALS in the same time frame\cite{Xie2008,Petrou2015,toth2016}. When working on locations of many hectares and larger, being able to automatically estimate the number of stems in the dominant canopy layers is a very useful tool and double-layer MCGC is able to provide this across several distinct forest types in the tropics.

\subsubsection{Biomass estimation}
\label{ssubsec:dis_MCGC_AGB}

Single-layer MCGC underestimated 1 ha plot-level carbon density (Figure \ref{fig:res_ACD_sgl}a). This is consistent with the under-detection of crowns discussed in Section \ref{ssubsec:dis_MCGC_FI}. The bias of -48\% shows that the estimates of biomass are about half of the values calculated based on field inventory, which is not unexpected. The algorithm finds more than half of the largest trees, which contribute the most biomass. It then finds decreasing numbers of the smaller trees and thus half the biomass is not detected. Applying forest specific correction factors removed most of the bias (Figure \ref{fig:res_ACD_sgl}b). However, given the initial estimates were heavily biased, we feel this is not an appropriate additional step. Two of the resulting correction factors take values of greater than 2, indicating a doubling of the initial estimate, which would produce large uncertainties in the estimates.

Double-layer MCGC, by contrast, produced overall estimates of plot carbon density with low bias (Figure \ref{fig:res_ACD_dbl}a). However, as shown by applying correction factors there is a tendency towards bias in differing forest types. In the original study of this data from Sepilok in [\citen{Coomes2017}] neither the ITC nor the area-based approach was able to produce estimates free of bias without local calibration or correction (-20\% and -19\% respectively). The results from double-layer MCGC do have a slightly larger RMSE of 33\% compared to 26\% and 20\% respectively in the original study. However, combining these with the bias of the methods suggests that double-layer MCGC produces more robust estimates of plot level carbon density than either methods in the original study. This is in contrast to the conclusion there that area-based methods remained the best option. Further, applying a forest type specific correction to the estimates did reduce the relative error to be similar to that for similar corrections in this original study (18\% compared to 18\% and 13\% respectively). However, for the ITC approaches we feel this correction isn't a fair reflection of the algorithm. This correction requires a local correction to be made and this would only be possible for areas where full field inventory are available and are unlikely to generalise to other regions or biomes. In a similar manner, we feel the fairest way to compare the area-based approach is with the general equation originally used in [\citen{Asner2014}]. Following the work in the original study on the data from Sepilok\cite{Coomes2017} this approach already requires local fitting of models for basal area and wood density, but the overall form of the equation is not locally fitted. In this comparison, ignoring locally fitted models or forest-type correction factors, double-layer MCGC produces the lowest bias overall across all 36 one ha plots (Table \ref{table:AGB}). Additionally, double-layer MCGC produces the lowest bias predictions of ACD for both Kerangas and Sandstone forest types (Table \ref{table:AGB_for}). Only in Alluvial forest is double-layer MCGC outperformed by the existing methods, though the potential difficulties in working in this forest type are discussed below. The overall best performing model was still the area-based modelling, once locally fitted. However, in a manner similar to applying a forest-type correction to ITC models, this approach requires local fitting of the overall equation for ACD, a local model for wood density and a choice of the best fitting model for predicting basal area as a function of gap fraction across a range of heights\cite{Asner2014,Coomes2017}. Thus the fairest and most direct comparison is to look at the output of the most general models, which are also the models most likely to generalise to other biomes and forests. As noted, in this comparison, double-layer MCGC performs best across all but the Alluvial forest type.

The reduced bias estimation of biomass is a powerful tool for the application of the MCGC individual crown detection algorithm. Our work used a regional allometric relationship (for the entire Indo-Malaya region) yet was able to produce initial estimates of biomass, before applying any correction factor, with lower bias than both the individual-tree region growing algorithm and area-based approaches used in [\citen{Coomes2017}]. It is worth noting that the overestimated Alluvial forest type has a particularly tall and unusual canopy structure\cite{Coomes2017}. Many trees in this forest type sit towards the extreme end of the data in [\citen{Jucker2017}]. It could therefore be advisable to develop a more specialised allometry for this forest type for use in the allometric feasibility checking step. Similarly any atypical forests in a given region may be better approached with a more specific allometric relation than the very general region-wide model used in this work. This would be further justified by this being the only forest type for which double-layer MCGC does not produce the best estimate of ACD across the method compared in Table \ref{table:AGB_for}.

\subsection{Possible extensions of MCGC}
\label{subsec:dis_ext}

double-layer MCGC is a very flexible approach to the problem of tree detection and biomass estimation. One obvious extension, which is in development, is to include spectral imagery data or other data which can be assigned to each point. This can easily be incorporated into equation \eqref{eq:w_Z} in a manner similar to the current terms. With a suitable dataset of an RGB or multispectral point cloud data combined with field inventory it should be possible to refine the delineation of crowns as part of MCGC. Similar modifications to account for return number or intensity are possible, all of which require modification only of the formula for linkage weights. The key strength of MCGC in this regard is the flexibility of the similarity computation. It is possible to alter the $w_{ij}$ term to reflect which factors are felt to be relevant in detecting crowns in any given forest.

Though double-layer MCGC still does not find all lower canopy trees, the balance of trees that it does find give rise to a good estimate of biomass at plot-level. These should combine well for analysis of UAV data, for which the lower layers of the canopy are often obscured and so do not appear in the structure from motion reconstruction. To be able to automatically produce an estimate of total plot biomass from UAV produced point clouds without any need for local calibration will be incredibly powerful for low cost, large sale project management. MCGC offers this capability, and bringing in a spectral element, as is present in UAV point clouds, should strengthen this further.

\section{Conclusion}
\label{sec:conc}

Our proposed double-layer MCGC approach to ITC detection and biomass prediction was able to identify the dominant trees of the canopy across three distinct tropical forest types. From the crowns of these trees MCGC was then able to produce a low-bias overall estimate of hectare-scale biomass and consequently carbon density in tropical forest. This is an advancement on previous work suggesting ITC approaches still need development to out-perform simpler area-based modelling. The strength of MCGC is its flexibility. The only constraints required to get these estimates of biomass are the four tuning parameters and a simple stem height to crown diameter allometry. For our work we used parameter values chosen by visual inspection, and we used a region-wide allometric relationship, which could be applied to any forest in the region. Hence MCGC was able to produce good estimates of biomass for a mixture of tropical forest structures, using only general parameters and knowledge. Similar approaches can be used to generate parameters and allometric relationships for any biome of interest. This is in contrast to area-based approaches which require either field measurements or local calibration, again requiring field measurements. MCGC permits local calibration through use of a local allometric relationship but work here shows the strength of working with a general regional allometry requiring no more than ALS data for the region of interest.

MCGC also incorporates multiple novel aspects which are not commonly included in tree crown detection algorithms. MCGC works directly on the point cloud, avoiding summarisation. MCGC also works with the raw point cloud, and does not first apply a ground height subtraction method, which skews trees on sloped ground. MCGC is also highly flexible and tunable through a small number of parameters whilst not being specific to any one forest type, though is able to use locally derived allometric relationships. All of this together makes MCGC a powerful tool in remote sensing of forests, supported by its strength in biomass estimation. MCGC also has potential for easy extension to multi-sensor datasets. Flexibility in the the construction of weights allows any combination of variables to be considered in detecting crowns and in particular this is well-suited to UAV-derived structure from motion data where the point cloud includes spectral data. A similar approach could be applied to the combination of hyperspectral imagery and LiDAR data once datasets have been aligned. As datasets combining multi-sensor imagery and field invetory become increasingly avaiable, MCGC can easily be extended and adapted to these, with only the construction of weights needing adjustment. 

\section{Acknowledgements}
\label{sec:acknow}

Jonathan Williams completed the work on this algorithm and its analysis as part of his PhD at the University of Cambridge, supported by NERC with RSPB as a CASE partner (NE/N008952/1). The project was further supported by a grant through the Human Modified Tropical Forests programme of NERC (NE/K016377/1). We thank members of the NERC Airborne Remote Sensing Facility and NERC Data Analysis Node for collecting and processing the data (project code MA14-14). David Coomes was supported by an International Academic Fellowship from the Leverhulme Trust. Oliver Phillips and Simon Lewis provided census data collected as part of an ERC Advanced Grant (T-Forces). Thanks to Richard Bryan Sebastian and Reuben Nilus for collecting field data on delineated trees and to Lindsay Banin for collecting data on field measured allometry. Xiaohao Cai was supported by the Issac Newton Trust and Welcome Trust. CBS acknowledges support from the Leverhulme Trust project on Breaking the non-convexity barrier, EPSRC grant Nr. EP/M00483X/1, the EPSRC Centre Nr. EP/N014588/1, the RISE projects CHiPS and NoMADS, the Cantab Capital Institute for the Mathematics of Information and the Alan Turing Institute. We gratefully acknowledge the support of NVIDIA Corporation with the donation of a Quadro P6000 GPU used for this research.


\bibliographystyle{unsrt}
\bibliography{MCGC}

\end{document}